\documentclass{article}

\usepackage{PRIMEarxiv}

\usepackage[utf8]{inputenc} 
\usepackage[T1]{fontenc}    
\usepackage{hyperref}       
\usepackage{url}            
\usepackage{booktabs}       
\usepackage{amsfonts}       
\usepackage{nicefrac}       
\usepackage{microtype}      
\usepackage{lipsum}
\usepackage{fancyhdr}       
\usepackage{graphicx}       
\graphicspath{{media/}}     
\usepackage{multirow}

\title{Large Language Models For Text Classification: Case Study And Comprehensive Review

}

\author{
  Arina Kostina, Marios D. Dikaiakos, Dimosthenis Stefanidis, George Pallis \\
  Computer Science, University of Cyprus \\
  Nicosia, Cyprus\\
  \texttt{akosti02@ucy.ac.cy, mdd@cs.ucy.ac.cy, dstefa02@ucy.ac.cy, gpallis@cs.ucy.ac.cy} \\
}

\begin{document}
\maketitle

\begin{abstract}
Unlocking the potential of Large Language Models (LLMs) in data classification represents a promising frontier in natural language processing. In this work, we evaluate the performance of different LLMs in comparison with state-of-the-art deep-learning and machine-learning models, in two different classification scenarios: i) the classification of employees’ working locations based on job reviews posted online (multiclass classification), and 2) the classification of news articles as fake or not (binary classification). Our analysis encompasses a diverse range of language models differentiating in size, quantization, and architecture. We explore the impact of alternative prompting techniques and evaluate the models based on the weighted F1-score. Also, we examine the trade-off between performance (F1-score) and time (inference response time) for each language model to provide a more nuanced understanding of each model’s practical applicability. Our work reveals significant variations in model responses based on the prompting strategies. We find that LLMs, particularly Llama3 and GPT-4, can outperform traditional methods in complex classification tasks, such as multiclass classification, though at the cost of longer inference times. In contrast, simpler ML models offer better performance-to-time trade-offs in simpler binary classification tasks.
\end{abstract}


\section{Introduction}

Text Classification is a fundamental task in natural language processing (NLP) that involves categorizing text into predefined labels or classes based on its content. This process is crucial for knowledge discovery and decision-making, as it allows us to extract valuable information from the data. As digital data continues to grow exponentially, the importance of effective text classification methods has become more prominent.

In parallel with this growth, advancements in deep learning and artificial intelligence have propelled Pre-Trained Language Models (PLM) to the forefront of NLP research. Among these PLMs, Large Language Models (LLMs) stand out for their impressive performance and significant promise in various tasks. Their deep learning architecture enables them to identify detailed and complex feature representations, which has led to substantial improvements in performance and adaptability in various NLP tasks, making them a focal point of interest in both - academic research and practical applications for data analysis ~\cite{Naveed2023ACO}.

The multilayered abilities of LLMs prompted us to investigate their effectiveness in addressing two complex classification tasks: (i) fake news detection, and (ii) employee review classification based on the working location. With these classification scenarios we compare multiple LLMs with ML and state-of-the-art models, and identify prompting strategies that can enhance LLM performance. In our analysis, we examine multiple dimensions, including models' performance, and time efficiency, exploring how model size, quantization, and prompting techniques can potentially impact the results. Specifically, we make the following contributions:

\begin{enumerate}
\item \textbf{Comparison of LLMs and traditional classification methods:} We provide a detailed evaluation of multiple LLMs, ML algorithms, and a state-of-the-art model on two text classification scenarios. This study seeks to determine whether the LLMs' ability to "understand" and process language contextually gives an advantage over traditional ML and state-of-the-art models in real-world classification challenges. Our analysis considers multiple aspects, using the performance-time trade-off, and weighted F1-score as key metrics.

\item \textbf{Identification of effective prompting techniques:} We analyze how different prompting strategies influence LLM performance, showing that upfront optimization can improve outcomes. We investigate the disparities in the influence of different prompting techniques on the models, analyzing the potential causes of differences in models' behavior.

\end{enumerate}
The findings of this investigation enhance our understanding of LLM capabilities and their applicability in real-world scenarios.

\section{Related Work}

Recent advances in deep learning have significantly transformed the field, introducing powerful models based on neural networks such as Convolutional Neural Networks (CNNs) ~\cite{Kim2014ConvolutionalNN,8622157,9378182}, Recurrent Neural Networks (RNNs) ~\cite{Wang2018JointEO}, and Transformer architectures ~\cite{Vaswani2017AttentionIA}. Models like GPT (Generative Pre-trained Transformer) ~\cite{Radford2018ImprovingLU}, BERT ~\cite{Devlin2019BERTPO}, and their variants, have profoundly impacted the field of NLP. The rapid advancements of LLMs have sparked numerous questions about their capabilities and applicability in real-world scenarios. Notably, there is a gap in the research in comparing and discussing LLM capabilities specifically in the task of text classification. An extensive overview of recent work in text classification using LLMs, covering a range of aspects, like models' training datasets, sizes, cost, and performance can be found in ~\cite{Fields2024ASO}. We build on this work by incorporating various prompting techniques to explore the full potential of LLMs in this task. Several studies cover details of the PML architectures and PLM-driven NLP techniques ~\cite{Min2021RecentAI}, while other cover a wide range of deep learning models ~\cite{Minaee2020DeepLT} or compare advanced PLMs with traditional methods ~\cite{REUSENS2024124302} in the task of text classification, including studies specifically targeting fake news detection ~\cite{khan2021benchmark}. However, these studies offer limited to no evaluation of LLMs and Transformer-based architectures. We expand the analysis by comparing traditional classification methods with LLMs and use a custom evaluation dataset for further assessment. Recent work in ~\cite{Zhang2023SentimentAI} discusses LLM performance improvements in sentiment analysis tasks, while also delving in prompt engineering analysis, comparing zero-shot and few-shot settings. Other studies focus on specific model families, like GPT, and evaluate their performance on a range of NLP tasks, such as Emotion Recognition Sentiment Analysis, and question-answering tasks ~\cite{KOCON2023101861,Chen2023HowRI}.

While many researchers experiment with LLMs to identify their most effective use cases, others focus on developing innovative techniques that enhance the LLM performance. Prompting techniques, for instance, have gained significant attention as a means to improve LLM performance without the need for down-stream fine-tuning. Researches have proposed different techniques, such as Chain-of-Thought (CoT) ~\cite{Wei2022ChainOT}, Tree-of-Thought - an extension of the CoT which explores multiple reasoning paths over thoughts ~\cite{Yao2023TreeOT}, Few-shot learning ~\cite{Brown2020LanguageMA}, Emotional Prompting ~\cite{Li2023LargeLM}, and other methods to obtain better performance. Recent research ~\cite{Sun2023TextCV} explores the potential LLMs in text classification tasks, introducing a prompting strategy that improves LLMs' reasoning abilities. A study in ~\cite{Sahoo2024ASS} offers a detailed overview of the prompting techniques and their specific applications. In contrast to these studies, our work provides an evaluation of various prompting techniques across multiple LLMs, specifically within the context of text classification tasks. We analyze how these techniques influence LLM reasoning and task understanding, providing insights into the potential causes of the performance variability across different prompts.

\section{Background on LLM and ML Models}
\subsection{Large Language Models}
Large Language Models (LLMs) are advanced artificial intelligence models, typically based on transformer architectures ~\cite{Vaswani2017AttentionIA}, trained on vast amounts of text data to understand and generate human-like text. LLMs have the capability to process and comprehend natural language in a wide range of contexts, enabling them to perform tasks such as language translation, text summarization, sentiment analysis, question answering, and more. On a broader scale, LLMs can be divided into 3 major categories based on their architecture:

\begin{enumerate}

\item \textbf{Encoder-only} models, like BERT ~\cite{Devlin2019BERTPO}, which consist only of transformer encoder layers. During the training, with the Masked Language Modeling (MLM) technique, some input tokens are replaced with a masking token. The encoder layers use self-attention to generate contextual embeddings for each token from both directions. Then, based on these embeddings of the surrounding text the model is asked to predict the original tokens. 

\item \textbf{Decoder-only} models, like GPT ~\cite{Radford2019LanguageMA}, which consist of transformer decoder layers. The layers use self-attention to process the input sequentially and predict the next tokens based on the previous ones. 

\item \textbf{Encoder-decoder} models, like T5 ~\cite{Raffel2019ExploringTL}. The encoder processes the entire input sequence and converts it into an encoded representation. Then the decoder takes this representation and reconstructs it step-by-step, using both the encoded context and previously generated tokens, into the output sequence.

\end{enumerate}

Models used for text generation, have key hyperparameters that can be tuned to control output quality, creativity, and coherence. These include 1) Temperature, which controls randomness — lower values (e.g., 0.2) make more predictable outputs, while higher values (e.g., 1.0 or above) increase creativity; 2) Top-k Sampling, which limits the model's word choices to the top-k most likely options; and 3) Max Tokens, which sets the maximum number of tokens the model can generate.
Moreover, as awareness of the ecological impacts of high computational resource use grows, techniques like quantization have emerged. Quantization techniques lower the precision of model weights and activations from floating-point (32-bit or 16-bit) to lower-bit formats (8-bit or 4-bit). This reduces model size, computation time, memory usage, and energy consumption. Some of the common quantization techniques are AWQ ~\cite{Lin2023AWQAW}, GPTQ ~\cite{Frantar2022GPTQAP}, QuaRot ~\cite{Ashkboos2024QuaRotO4}, Atom ~\cite{Zhao2023AtomLQ}.

In our experiments, we compare decoder-only models with RoBERTa, an encoder-only baseline for text classification, as encoder-decoder models are conventionally used for sequence generation tasks. The LLMs used for the experiments differed in several key aspects: model size, foundational architecture, quantization status, and deployment method (local installation versus API-accessible services such as Groq and GPT). Specifically, the models for our experiments are the following:

\subsubsection{Decoder-Only Models}

\begin{enumerate}
\item Mistral-7B OpenOrca (\textbf{Mistral-OO}). Mistral-OO is an instruct-tuned Mistral-7B model ~\cite{Jiang2023Mistral7}, trained on the dataset reproduced from the Microsoft Research team's OpenOrca Paper ~\cite{Mukherjee2023OrcaPL,longpre2023flan,lian2023mistralorca1}. 

\item OpenHermes 2.5 Mistral-7B (\textbf{Mistral-OH}). Mistral-OH is a fine-tuned Mistral-7B model, trained on GPT-4 generated data and various open datasets ~\cite{OpenHermesMistral}. 

\item zephyr-7B-beta (\textbf{Zephyr}). Zephyr is a fine-tuned, intent-aligned version of Mistral-7B, meaning that it has an improved ability to grasp onto the user’s preference and intent. To achieve this objective, the methodology involved  Direct Preference Optimization (DPO) training ~\cite{rafailov2024direct} and additional fine-tuning of the Mistral-7B model ~\cite{Tunstall2023ZephyrDD}. 

\item Nous-Hermes Llama2 13B (\textbf{Llama2}). Llama 2 ~\cite{Touvron2023Llama2O} is a family of pre-trained LLM models that were developed by researchers at Meta. The particular model is additionally fine-tuned on high-quality GPT-4 datasets ~\cite{Llama2Hermes}.

\item Xwin-MLewd 13B v0.2 (\textbf{Xwin}). Xwin utilizes the Llama2 model as its foundation, which was additionally exposed to various training data and underwent Reinforcement Learning (RL) training ~\cite{kaelbling1996reinforcement} ~\cite{xwin-lm}. 

\item Gemma 2 9B (\textbf{Gemma}). Gemma was developed by Google as a family of lightweight, state-of-the-art open models, suitable for environments with limited resources due to their small size ~\cite{gemma_2024}.

\item Meta Llama 3 70B (\textbf{Llama 3 70B}); and

\item Meta Llama 3 8B (\textbf{Llama 3 8B}) are both part of the Llama 3 family models, which are auto-regressive language models developed by Meta that use an optimized transformer architecture ~\cite{llama3modelcard}. 

\item Mixtral 8x7B (\textbf{Mistral}). The Mixtral-8x7B Large Language Model (LLM) is an open-source, state-of-the-art language model ~\cite{Jiang2023Mistral7}.

\item gpt-4-turbo (\textbf{Gpt4-turbo}). The Gpt4-turbo is a high-intelligence model developed by the OpenAI ~\cite{gpt4turbo}.

\end{enumerate}

\subsubsection{Encoder-Only Model}
In this experiment, we utilized the state-of-the-art model RoBERTa (Robustly optimized BERT approach) ~\cite{Liu2019RoBERTaAR}, which builds upon BERT's architecture ~\cite{Devlin2019BERTPO} with modifications to training objectives, hyperparameters, and training data. The implementation details and the hyperparameters used for the trainings are discussed further in section \ref{sec:config}

\subsubsection{Prompt Engineering}
\begin{figure*}
\centering
\input{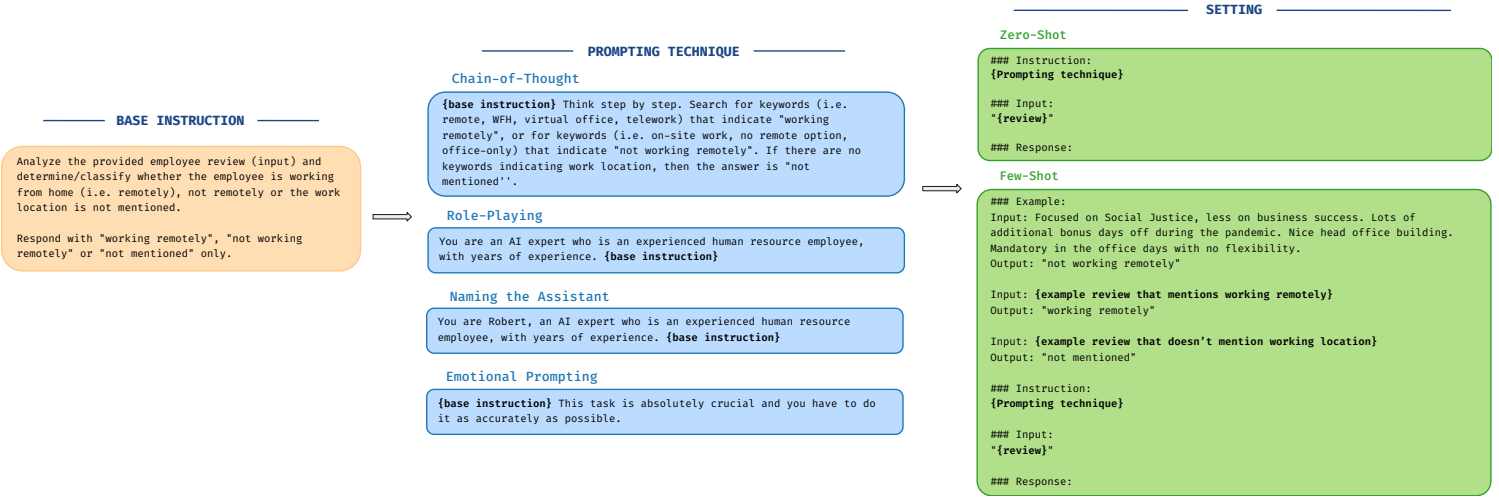}


\caption{\small Example Prompt Construction: The base instruction combines with a prompting technique, then is wrapped in a ZS or FS setting, which forms the final prompt sent to the LLM.}

\label{fig:Poster Prompts}
\end{figure*}

Prompt engineering is a technique that involves crafting and phrasing instructions in a way that guides the LLMs toward producnpng the desired outputs ~\cite{Bsharat2023PrincipledIA}. We construct our prompts in Zero-Shot and Few-shot settings, where: 

\begin{enumerate}
\item \textbf{Zero-shot (ZS)}, is a method that gives only the base instruction of the task to the model with no additional attempt to evoke reasoning, so the model relies solely on its pre-existing knowledge to produce a response ~\cite{Radford2019LanguageMA}.

\item \textbf{Few-shot (FS)}, is a method that apart from the the base instruction, additionally provides the model with a small number of training examples ~\cite{Brown2020LanguageMA}. 
\end{enumerate}
We used the following prompt engineering techniques: 

\begin{enumerate}
\item \textbf{Chain of Thought (CoT)}, a technique that explains the logic and the reasoning behind a correct response, providing a step-by-step thought process to the model ~\cite{Wei2022ChainOT}.

\item \textbf{Emotional prompting (EP)}, which is based on appending a psychological phrase to the end of an existing prompt, that acts as emotional stimuli for the LLM ~\cite{Li2023LargeLM}.

\item \textbf{Role-playing (RP)}, also referred to as the “Persona” theme, a strategy that provides the persona’s description in the prompt to make the LLMs adopt a specific character and take on specific perspectives to solve the given task ~\cite{Bsharat2023PrincipledIA}.

\item \textbf{Naming the Assistant (NA)}, a practice of giving a nickname to the LLM. In our experiment, this technique is used in combination with the Role-playing technique by appending the name to the beginning of the instruction.

\end{enumerate}

All prompting techniques are constructed upon the same base instruction. This ensures that we have an objective evaluation of how each one of the prompting techniques affects the output.

\subsection{Traditional ML Models}
For our experiments we use 2 Machine Learning (ML) models: \\
\textbf{Naive Bayes (NB)}, which is an ML algorithm based on Bayes' theorem, commonly used in classification tasks, and \\
\textbf{Support Vector Machines (SVM)}, which is a supervised ML algorithm that is often used in classification tasks, due to its ability to handle high-dimensional data effectively ~\cite{Cortes1995SupportVectorN}. \\
Further implementation details are also discussed in section \ref{sec:config}.

\section{Experimental Setup}
\subsection{Configurations and Implementation} \label{sec:config}

\textbf{Hardware:} The Groq and OpenAI models were accessed with an API key, with their computational resources managed externally. Specifically, Groq models (Mistral, Llama3 70B, Llama3 8B, Gemma) run on the Groq Language Processing Unit (LPU), hardware designed for AI inference ~\cite{GroqLPU}. For Gpt4-turbo, details on model architecture, and hardware are not disclosed by OpenAI. The five open-source AWQ-quantized ~\cite{Lin2023AWQAW} models (Llama2, Xwin, Mistral-OO, Mistral-OH, Zephyr) were sourced from HuggingFace ~\cite{wolf-etal-2020-transformers} - a platform that provides an open-source library of a wide range of pre-trained models - and loaded on NVIDIA Tesla T4 GPU in Google Colaboratory ~\cite{Colab}. RoBERTa and ML algorithms were also trained on the T4 GPU in Google Colaboratory. \\
\textbf{LLM Hyperparameters:} To eliminate randomness, and ensure precise responses solely based on the input, the temperature for the models was set to 0. We request outputs only once per prompt, as a previous small-scale experiment showed that with temperature 0, repeating the questions consistently gave identical results. \\
\textbf{RoBERTa Training:} We use Transformers library ~\cite{Wolf2019TransformersSN} to access pre-trained models and PyTorch library ~\cite{Paszke2019PyTorchAI} for fine-tuning. The process involved loading the pre-trained RoBERTa model and modifying the output layer to match our specific classification task. Then, we trained the model on our annotated dataset, using the Adam ~\cite{Kingma2014AdamAM} optimizer, which adapts learning rates for each parameter. We apply 5-fold cross-validation, and set the learning rate to 1e-5 and a batch size to 32, which are commonly used values that balance computational efficiency and learning stability. \\
\textbf{NB and SVM Training:} We use the scikit-learn library ~\cite{Pedregosa2011ScikitlearnML}, which provides the NB and SVM classifiers, different ML tools, and text feature extraction methods. For each model our pipeline utilizes TfidfVectorizer ~\cite{Pedregosa2011ScikitlearnML}, optimized via GridSearchCV ~\cite{Pedregosa2011ScikitlearnML}, which uses grid search with 5-fold cross-validation to choose the optimal combinations of parameters. Feature extraction varies by dataset and model, aiming to optimize model performance for each task: FakeNewsNet uses up to 5000 (for SVM) and 10000 (for NB) features, while Employee Reviews uses up to 10000 (for SVM) and 5000 (for NB) features. \\

\textbf{LLM Querying:} For the HuggingFace models we use the vLLM library, which improves the throughput by batching multiple requests together ~\cite{Kwon2023EfficientMM}. Groq does not have an inference for request batching, so the prompts were sent sequentially. The free Groq plan comes with certain API request rate limitations (e.g. Llama3 70B allows 6,000 tokens per minute), which introduced additional delay in obtaining the results. GPT4-turbo model was accessed with batched requests, which optimized the inference costs and handled rate-limiting constraints, which is ideal for high-volume dataset classification.

\subsection{Datasets} \label{dataSection}
Our experiments focus on 2 classification scenarios: binary and multiclass classification (3 classes), for which we utilize two distinct datasets. To select the samples that form our datasets we excluded records exceeding 4,096 tokens, to ensure compatibility with all the models' input constraints, and the records that we reserved as examples for the Few-Shot setting.
The same datasets were used to train the RoBERTa model, and the NB and SVM classifiers.

\subsubsection{FakeNewsNet Dataset} This is a widely recognized and publicly available dataset ~\cite{shu2018fakenewsnet,shu2017fake,shu2017exploiting} used for a binary classification task, where the model must categorize each article as either "fake" or "real". The challenge lies in the model's ability to detect subtle clues such as sensationalist language, exaggerated claims, or unverifiable facts, which helps to distinguish real news from fake. Additionally, fake articles may selectively use facts alongside misleading or false information, complicating the classification process. For our analysis, we selected a total of 214 reviews from the  '\verb|politifact_fake.csv|' and '\verb|politifact_real.csv|' files.

\subsubsection{Employee Reviews Dataset}
This dataset consists of 1,000 employee company reviews sourced from a platform where current and former employees provide anonymous feedback on companies and their management. To ensure balanced classes, we first filtered a larger pool of reviews using location-related keywords, then, based on the keyword groupings we randomly selected an equal amount (\textasciitilde33\%) of reviews for each class, and manually annotated the reviews into the categories. This presents a multi-class classification task, where the model categorizes the reviews into three classes: "working remotely," "not working remotely," or "not mentioned". This task presents an interesting challenge due to its real-world context, the growing relevance of remote work in modern work environments, and increased complexity compared to binary tasks.

\begin{table}[ht]
\centering
\caption{Datasets Details}

\label{tab:reviews-dataset}
\setlength{\tabcolsep}{2.5pt} 
\begin{tabular}{l|c|c}
\hline
\textbf{Metric} & \textbf{Employee Reviews} & \textbf{FakeNewsNet} \\

\hline
Total Records & 1,000 & 214 \\
\hline

& Working Remotely 37\% (372) & Fake 50\% (107) \\
Classes & Not Working Remotely 28\% (279) & Real 50\% (107) \\
& Not Mentioned 35\% (349) & \\

\hline
Avg Length & 108 tokens & 593 tokens \\
\hline
\end{tabular}


\end{table}

\subsection{Evaluation Metrics}
\subsubsection{Weighted F1-score}

We use weighted F1-score as our main metric, which is a balanced measure between precision and recall, and also considers the proportion for each class, giving more weight to classes with more instances. This ensures proportional performance evaluation across all classes, crucial in our experiment due to a slight class imbalance in the Employee Reviews dataset. For the RoBERTa and ML models, as the datasets are relatively small (214 and 1000 records), we utilize k-fold cross-validation, mitigating the risk of overfitting that can occur with small datasets, and use the mean weighted F1-score across folds as our primary metric.

\subsubsection{Time}
We consider the F1 score - time trade-off from the user’s perspective. We measure the total response time for Hugging Face’s and API-accessed LLMs, and inference time on Google Colab's T4 GPU for the RoBERTa and traditional ML models.

\section{Result Analysis}
\subsection{Model Comparison}

\begin{table*}[htbp]
\centering
\resizebox{\textwidth}{!}{%
\begin{tabular}{l|c c c c c c c c c c c c c}
\hline
\textbf{Prompting Method} & \textbf{Llama3 70B} & \textbf{Llama3 8B} & \textbf{Gemma2} & \textbf{Mistral} & \textbf{Mistral OO} & \textbf{Mistral OH} & \textbf{Zephyr} & \textbf{Llama2} & \textbf{Xwin} & \textbf{Gpt4-turbo} & \textbf{RoBERTa} & \textbf{Naive Bayes} & \textbf{SVM} \\
\hline
\multicolumn{14}{c}{\textit{FakeNewsNet}} \\
\hline
1) ZS & 92.5 & 89.2 & 82.8 & 84.5 & 83.1 & 81.6 & 76.8 & 78.3 & 47.9 & 81.7 & \multirow{10}{*}{93.0} & \multirow{10}{*}{90.0} & \multirow{10}{*}{88.8} \\
2) ZS+COT & 91.1 & 78.4 & 80.4 & 80.8 & 83.6 & 84.6 & 84.6 & 77.8 & 78.0 & 81.7 & & & \\
3) ZS+EM & 92.0 & 89.2 & 83.3 & 84.1 & 84.1 & 79.1 & 75.9 & 79.4 & 45.9 & 83.2 & & & \\
4) ZS+RP+NA & \textbf{\textit{94.4}} & 88.3 & 82.8 & 84.1 & 84.5 & 78.8 & 78.0 & 53.3 & 48.7 & 82.2 & & & \\
5) ZS+COT+EM & 91.1 & 86.4 & 82.3 & 86.0 & 83.6 & 85.5 & 85.0 & 80.8 & 76.2 & 81.7 & & & \\
6) ZS+RP & 93.0 & 87.4 & 83.3 & 83.2 & 83.5 & 77.4 & 75.4 & 47.8 & 46.9 & 82.7 & & & \\
7) FS & 92.5 & 86.4 & 81.4 & 80.8 & 81.2 & 81.0 & 75.6 & 52.1 & 60.9 & 82.3 & & & \\
8) FS+RP+NA & 92.5 & 86.0 & 82.8 & 85.5 & 79.7 & 80.7 & 71.2 & 41.1 & 55.8 & 83.7 & & & \\
9) ZS+COT+RP+NA & 93.0 & 80.8 & 81.3 & 85.0 & 83.6 & 82.2 & 79.9 & 74.8 & 70.1 & 82.2 & & & \\
10) FS+COT+RP+NA & 93.6 & 83.2 & 81.7 & 87.8 & 81.9 & 81.4 & 81.1 & 44.1 & 69.1 & 82.8 & & & \\
\hline
\multicolumn{14}{c}{\textit{Employee Reviews}} \\
\hline
1) ZS & 83.9 & 81.9 & 82.8 & 83.3 & 83.4 & 71.6 & 61.1 & 50.7 & 52.2 & 80.2 & \multirow{10}{*}{83.8} & \multirow{10}{*}{61.3} & \multirow{10}{*}{68.7} \\
2) ZS+COT & 86.9 & 83.7 & 80.4 & 81.4 & 85.8 & 81.9 & 76.3 & 53.0 & 64.1 & 85.2 & & & \\
3) ZS+EM & 83.4 & 82.0 & 83.3 & 83.4 & 84.4 & 72.9 & 62.2 & 47.8 & 55.7 & 81.6 & & & \\
4) ZS+RP+NA & 81.7 & 81.0 & 82.8 & 82.5 & 81.2 & 70.9 & 60.3 & 52.6 & 56.5 & 81.8 & & & \\
5) ZS+COT+EM & 86.9 & 84.0 & 82.3 & 81.1 & 86.4 & 81.1 & 75.7 & 48.4 & 62.3 & 85.6 & & & \\
6) ZS+RP & 80.7 & 82.1 & 83.3 & 82.1 & 81.0 & 72.6 & 60.1 & 52.3 & 52.0 & 82.5 & & & \\
7) FS & 84.1 & 81.8 & 84.2 & 80.9 & 80.4 & 75.0 & 70.5 & 55.8 & 70.7 & 84.7 & & & \\
8) FS+RP+NA & 82.2 & 82.1 & 84.1 & 81.3 & 80.0 & 72.8 & 71.2 & 52.7 & 71.0 & 85.7 & & & \\
9) ZS+COT+RP+NA & 87.1 & 83.2 & 81.5 & 80.7 & 85.9 & 81.1 & 75.9 & 54.0 & 64.3 & 84.8 & & & \\
10) FS+COT+RP+NA & 86.8 & 84.9 & 84.7 & 82.5 & 81.7 & 78.5 & 71.7 & 62.7 & 74.2 & \textbf{\textit{87.6}} & & & \\
\hline
\end{tabular}
}

\caption{F1-Score Results of the Models Across Various Prompting Methods}
\label{tab:prompting_comparison}
\end{table*}

We observe that for the FakeNewsNet Dataset Llama3 70B  achieved the highest F1-score of \textbf{\textit{94.4\%}} using the ZS+RP+NA prompt, with its other F1 scores consistently being above 91\% across all the prompts. RoBERTa achieved an excellent score of 93.0\%, surpassing all the other LLMs and ML models. This peak performance was observed after 7 epochs of training, based on the empirical evaluation over 10 epochs. The ML models NB and SVM also demonstrated competitive performance, achieving 90.0\% and 88.8\% F1 scores respectively, surpassing all of the LLMs except Llama3 70B and 8B, and RoBERta. Surprisingly, the Gpt4-turbo model, which typically excels in various tasks, was outperformed by 5 models, and struggled on this dataset with its best score being 83.7\% using FS+RP+NA prompt. Zephyr, Xwin, and Llama2 showed variable performance, with the first two achieving mid-80s F1 scores, while Llama2 had the worst overall performance.

\begin{figure}[t]
  \centering
     \includegraphics[width=0.7\textwidth]{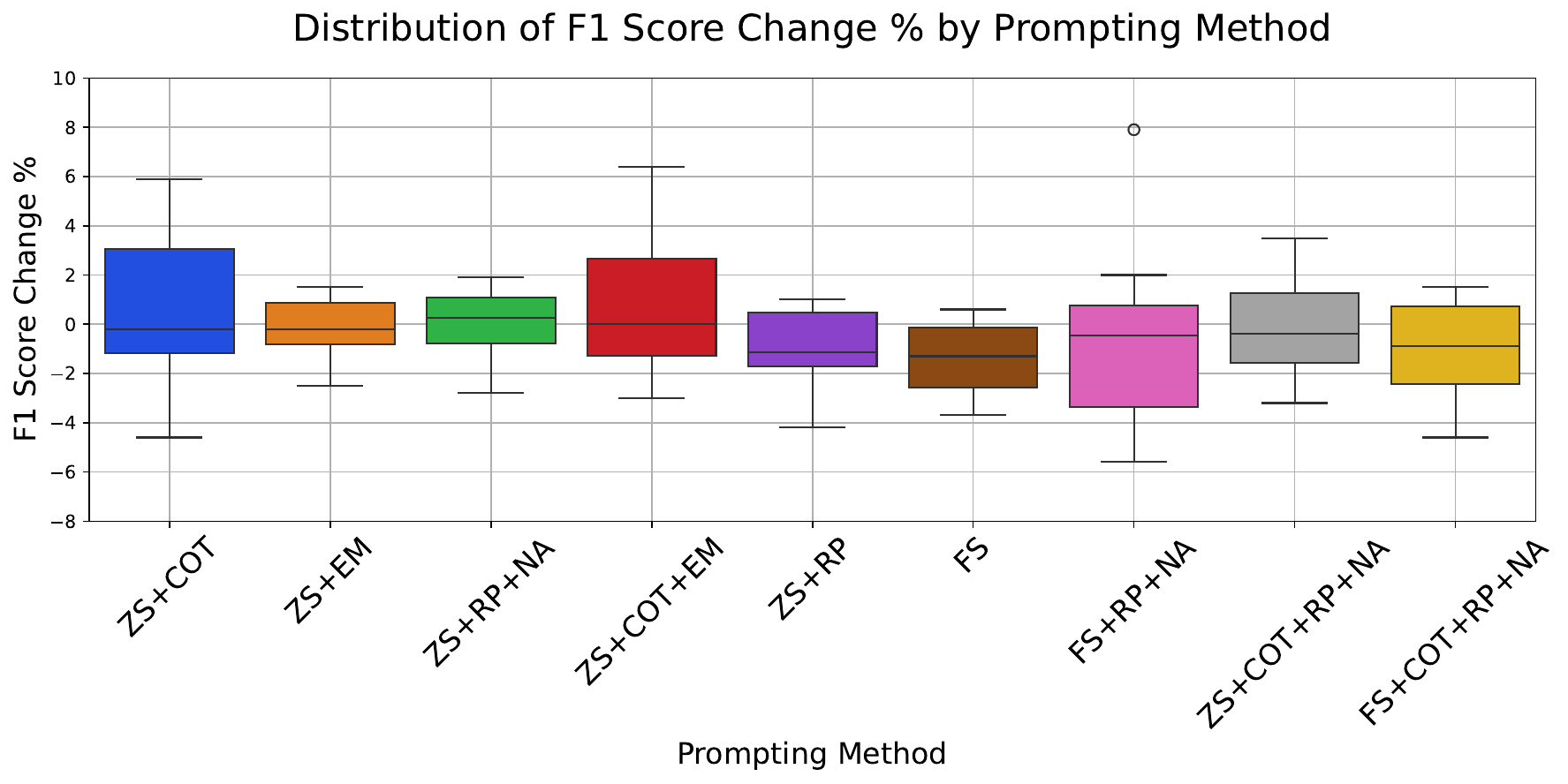}

    \caption{Boxplot of F1 Score \% change that Each Prompt Caused Compared to Basic ZS (FakeNewsNet Dataset)}
    
  \label{fig:All_fakenews_boxplot_final}
  
\end{figure}
\begin{figure}[t]
  \centering
     \includegraphics[width=0.7\textwidth]{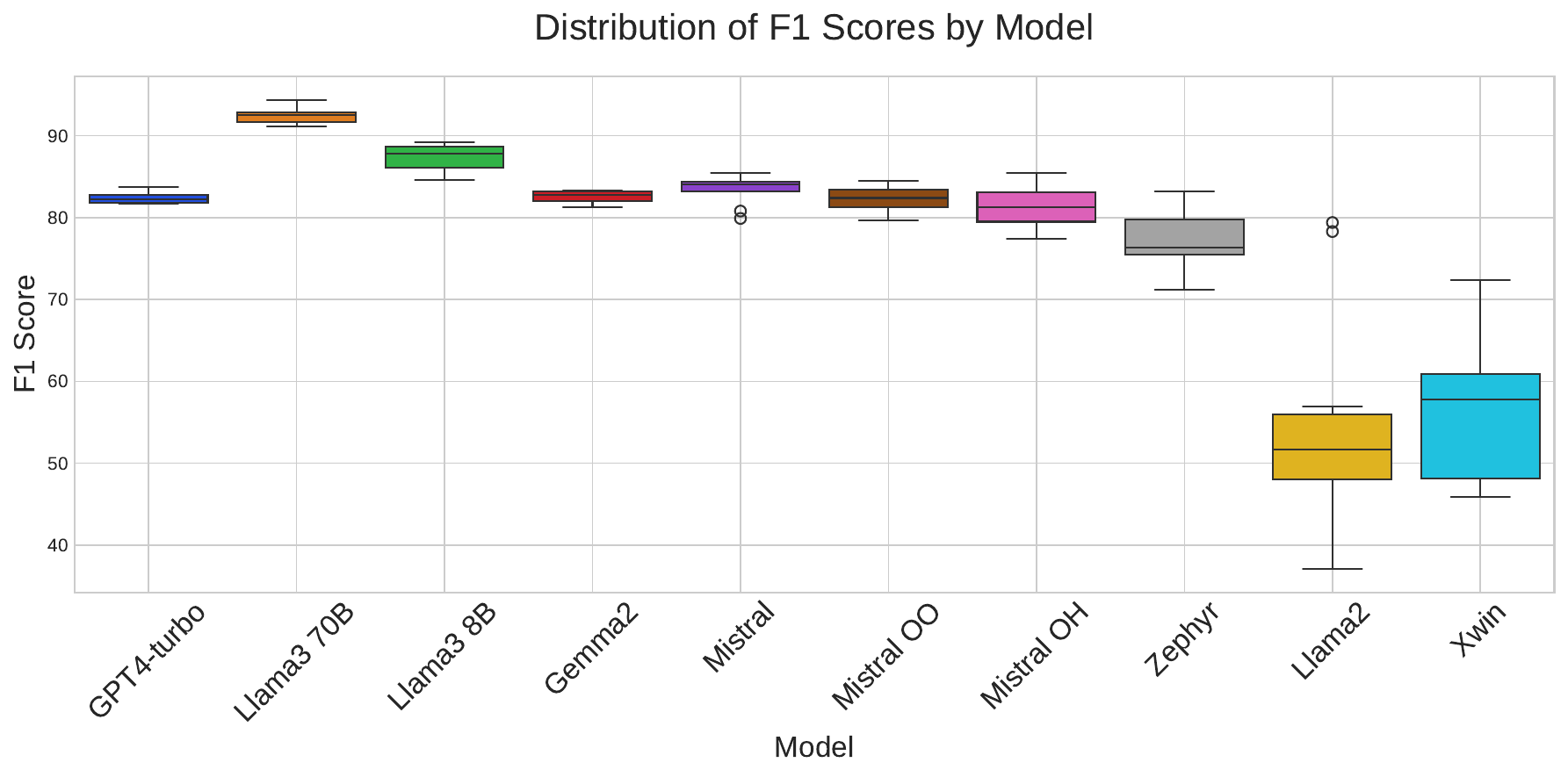}

    \caption{Boxplot of Performance Range for Each Model (FakeNewsNet Dataset)}

      \vspace*{-1\baselineskip}
    
  \label{fig:All_FakeNewsNet_model_boxplot_final}
  \vspace*{-1\baselineskip}
  
\end{figure}

For the 3-class classification task, Gpt4-turbo achieved the highest F1-score of \textbf{\textit{87.6\%}} using the FS+COT+RP+NA prompting method, with Llama3 70B and Mistral OO, falling behind by only 0.5\% and 1.2\% respectively. Llama3 8B achieved the 2nd highest score, reaching 84.9\% with FS+COT+RP+NA prompt. RoBERTa maintained a strong performance with an F1-score of 83.8\% achieved after 5 epochs of training, although it is surpassed by 5 LLM models, being closely behind with the highest score difference being 3.7\%. Gemma2, similarly to Gpt4-turbo showed improved performance compared to the FakeNewsNet task, with the F1-score reaching 84.9\%. Zephyr, Llama2, and Xwin again showed variable performance, with the last two particularly struggling on this task.

\subsection{Impact of Prompting methods}

\begin{figure}[t]
  \centering
     \includegraphics[width=0.7\textwidth]{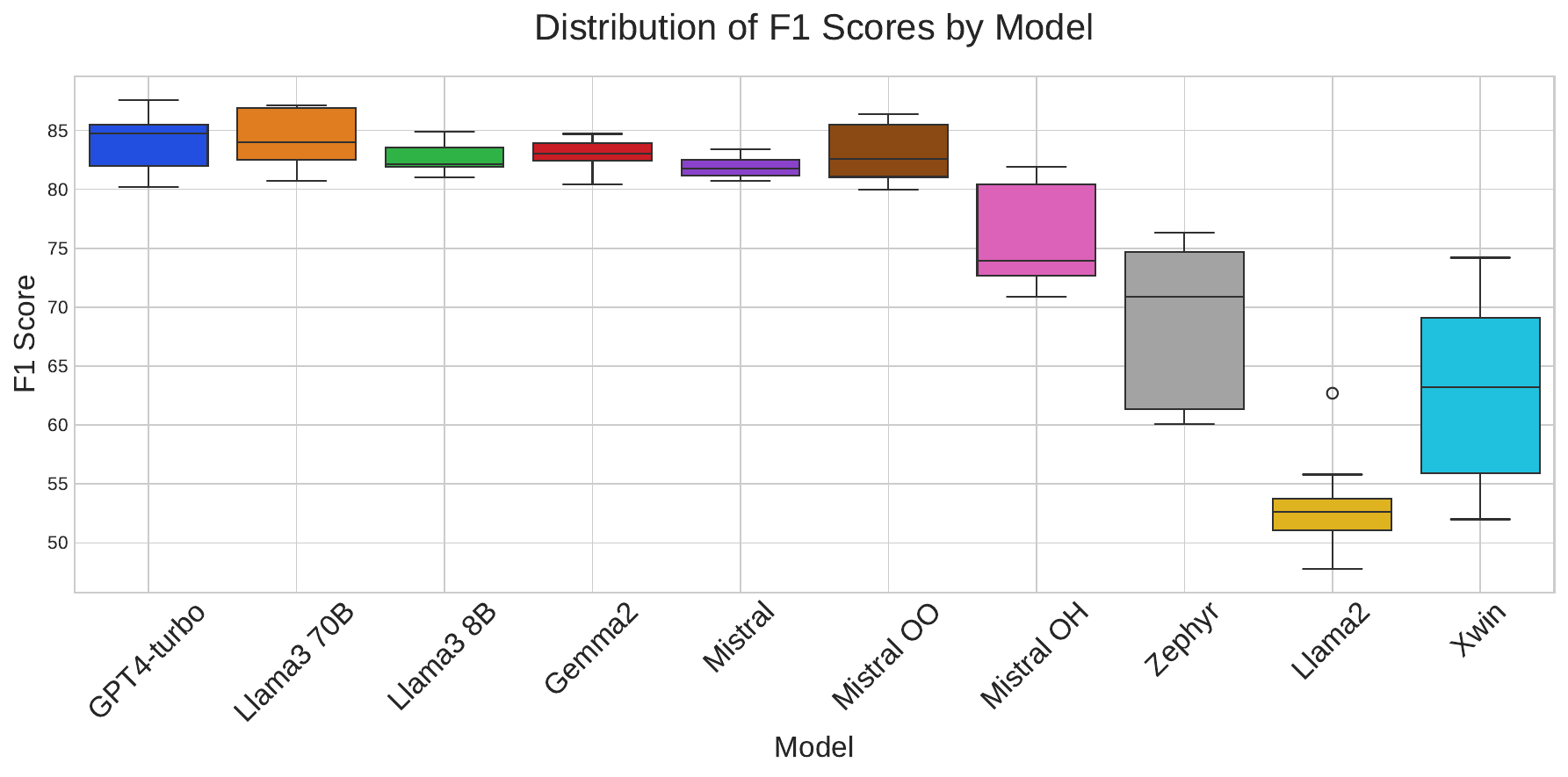}
          

    \caption{Boxplot of Performance Range for Each Model (Employee Reviews Dataset)}
    
  \label{fig:All_Employee_Reviews_model_boxplot_final}
  
\end{figure}
\begin{figure}[t]
  \centering
     \includegraphics[width=0.7\textwidth]{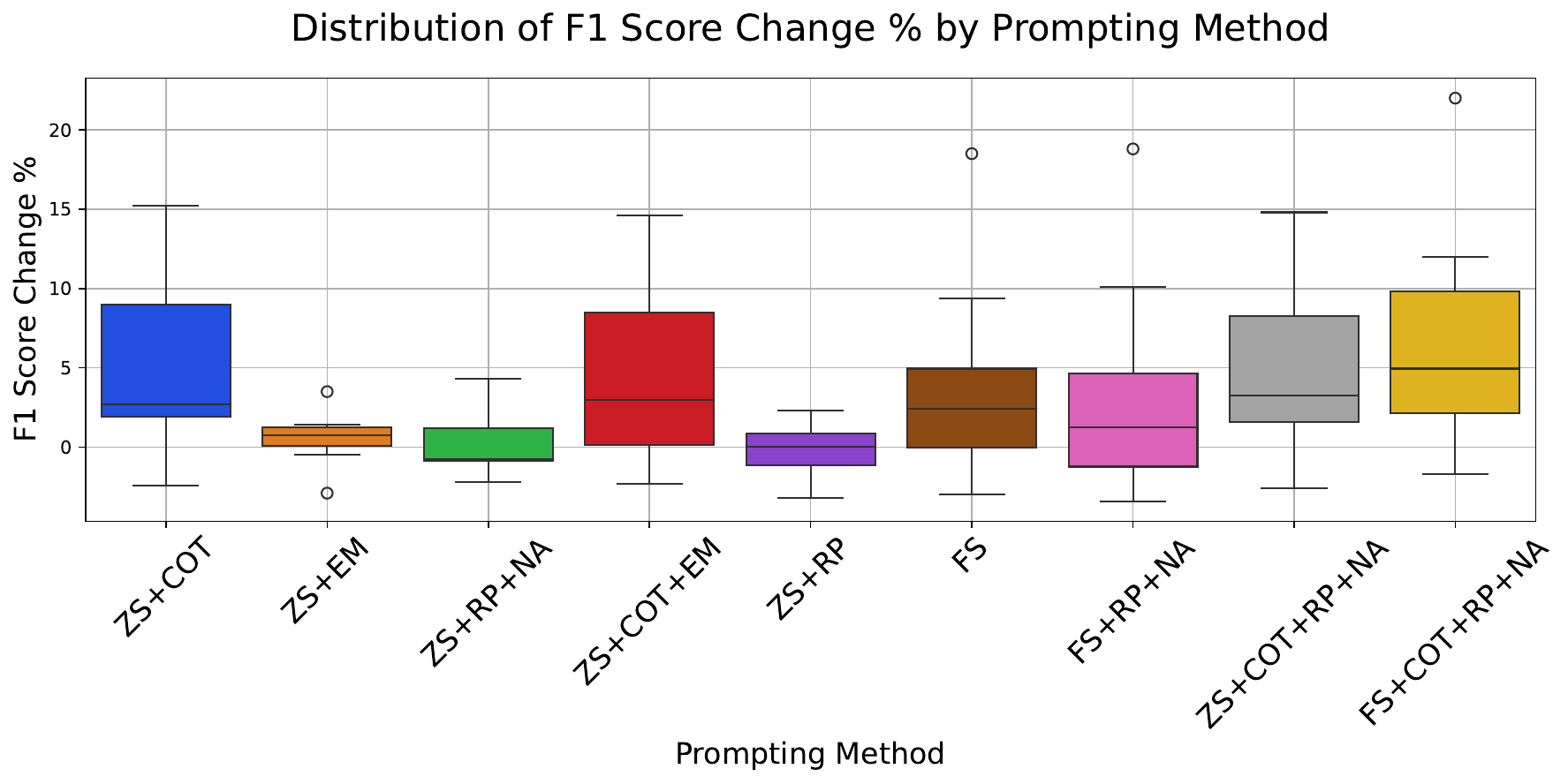}
          

    \caption{Boxplot of F1 Score \% change that Each Prompt Caused Compared to Basic ZS (Employee Reviews Dataset)}
    
  \label{fig:All_Employee_Reviews_boxplot_final}
  
      \vspace*{-1\baselineskip}
\end{figure}

For the Employee Reviews dataset, from Figure \ref{fig:All_Employee_Reviews_boxplot_final} we can see that by utilizing different prompting techniques, the models can achieve a much higher F1 score, reaching as much as a \textbf{\textit{22.2\%}} point increase compared with the basic ZS approach. Specifically, the results demonstrate that CoT is a powerful technique that can improve performance in both - ZS and FS settings. Also, the FS setting, particularly when combined with other prompting techniques, demonstrated a reasonable performance increase across most models, showing that the examples provide additional information and context for the task ~\cite{Brown2020LanguageMA}. For the particular task, the three best prompting techniques appeared to be FS+COT+RP+NA, ZS+COT+RP+NA, and ZS+COT+EM. We observe that the latter method - where we combine multiple techniques, yields better results. However, prompts employing the Naming-the-Assistant (NA) and Role-playing (RP) techniques often reduced performance, resulting in an F1-score inferior even to that of the basic ZS approach. A possible explanation could be the additional challenge of providing accurate results while trying to align with the given persona.

In Figure \ref{fig:All_fakenews_boxplot_final} for the FakeNewsNet dataset, we observe a different behavior compared to the previous classification scenario, with the best techniques being ZS+RP+NA, standard ZS, and ZS+COT+EM. Surprisingly, while most of the prompting techniques led to a decreased performance, the prompt ZS+RP+NA employing the Role-Playing and Naming-the-Assistant techniques, appeared to be the most stable, providing a relatively small but consistent performance increase to most of the models. Also, similarly to the Employee Reviews dataset, prompts employing the CoT technique managed to yield higher performance in some cases (ZS+COT+EM, ZS+COT).

This behavior difference can be attributed to the ambiguity in how LLMs interpret natural language. LLMs may find certain information in prompts irrelevant and distracting, even if it appears essential to humans, which may cause a drop in performance ~\cite{Shi2023LargeLM}. This is largely because there is no established theoretical explanation of how LLMs interpret natural language, which makes their internal decision-making process somewhat opaque ~\cite{Chang2024EfficientPM}. Table \ref{tab:prompting_comparison} shows differences in model performance across the prompting techniques, indicating that prompts are not universally effective. While some models benefit from specific phrasing, others do not, emphasizing the lack of generalizability in prompt design across different LLMs ~\cite{Zhao2021CalibrateBU,Schick2020ExploitingCF,Perez2021TrueFL}.

\subsection{Variability In Performance}
From figures \ref{fig:All_FakeNewsNet_model_boxplot_final}, \ref{fig:All_Employee_Reviews_model_boxplot_final} we can observe the variability of the model performance. This is expected since, despite extensive training, LLMs most certainly have not seen every conceivable phrasing of a certain task. Consequently, the model is required to speculate on the possible tasks and answer that the user expects to get, relying on semantic similarity with questions in its training data. Hence, the significance of prompt engineering is evident, as a poorly constructed prompt can potentially mislead the model, resulting in poor performance ~\cite{Jiang2019HowCW}. 
Interestingly, models like Xwin, Llama2, and Zephyr had the biggest fluctuations in the results, while also having the worst median performance among all of the models. Llama2 had the biggest performance variability with the FakeNewsNet dataset, with a significant \textbf{\textit{42.3\%}} difference between its best and worst results, while Xwin, with the Employee Reviews dataset, had the biggest performance difference of \textbf{\textit{22.2\%}}. This behavior suggests a correlation between the performance of a model and its sensitivity to the phrasing of the task. Models with lower performance tend to have bigger fluctuations across different prompts. Conversely, higher-performing models, which are more certain about their answers, are less affected by the phrasing of the task.

\subsection{Model Scaling and Quantization}

Our results reveal that Llama3 70B dominates the leaderboard, consistently outperforming all of the smaller models. In contrast, smaller models such as Llama3 8B and Mistral exhibit decent performance, especially on the Employee Reviews dataset. Although they trail Llama3 70B by only 2.2\% and 0.7\% respectively on the Employee Reviews dataset, they still cannot match the results of larger models, even with advanced prompting. We observe that the quantized Mistral models manage to match, and even exceed the performance of the standard Mistral model. Notably, Mistral-OO outperforms the standard Mistral by \textbf{\textit{4.5\%}} on the Employee Reviews dataset. This observation reveals two insights: that quantization did not negatively impact Mistral's efficiency; and that the additional fine-tuning improved its performance, highlighting the importance of a carefully-chosen training dataset. While quantized Mistral models can match and surpass their non-quantized version, we cannot generalize it to all models. For instance, the notable performance gap between Llama2-based models (Llama2, Xwin), and Llama3-based models may result from their different architectures and not from the application of quantization.

\begin{figure}
\hspace*{-1.5cm}
\includegraphics[width=.6\linewidth]{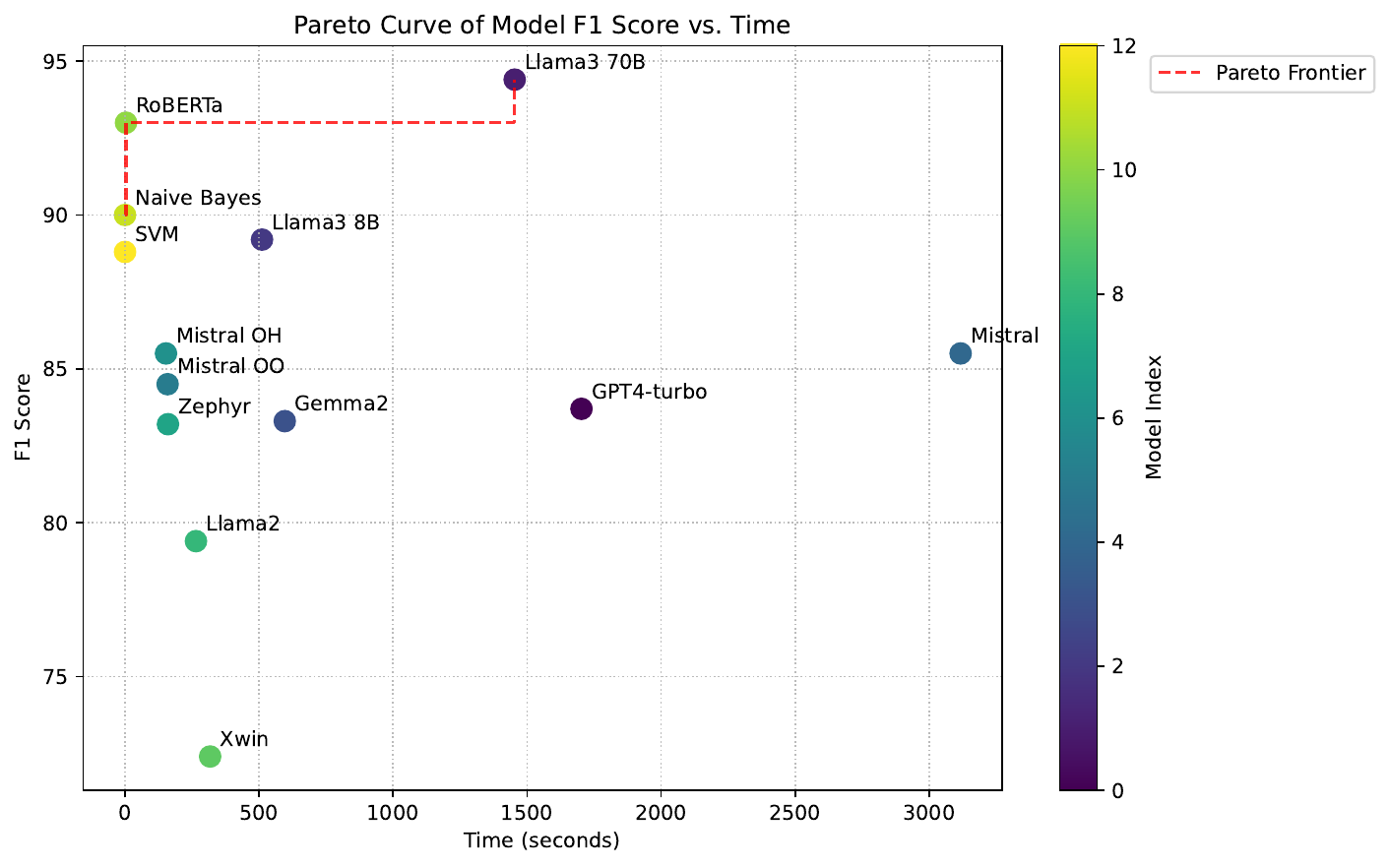}
\hfill
\includegraphics[width=.6\linewidth]{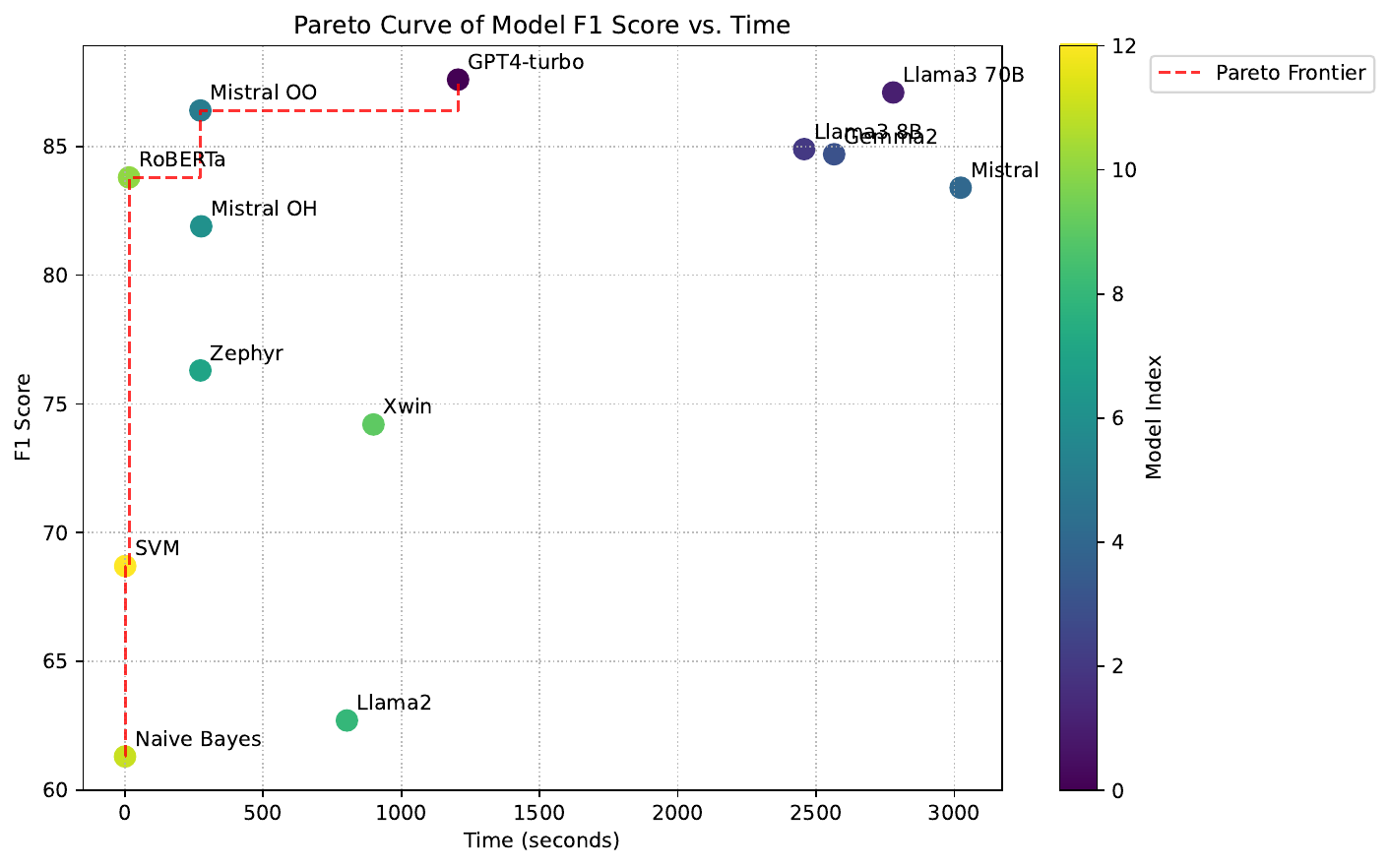}

\vspace*{-0.7\baselineskip}

\caption{Pareto Curve for F1-score - time trade-off for the Employee Reviews (right) and FakeNewsNet (left) Datasets}
\label{fig:paretro-curves}
\vspace*{-1\baselineskip}

\end{figure}

\subsection{Trade-offs and Practical Implications}
\vspace*{-0.5\baselineskip}
The Pareto curves in Figure \ref{fig:paretro-curves} illustrate the trade-off between F1-score and time required to obtain results. Pareto curves visually represent the trade-off between two competing objectives, where improving the one often comes at a cost of the other. This allows to choose the appropriate model, that offers the best performance for a reasonable time, or conversely, requires the least amount of time to achive the desired level of performance. For the Employee Reviews dataset, Gpt4-turbo exhibits the highest F1-score (\textasciitilde88\%), but at a significant time cost (over 2500 seconds). Mistral OO, while performed slightly worse, required less than 300 seconds. RoBERTa scored a competitive \textasciitilde84\% in only 15 seconds. The traditional models SVM and NB offer the lowest F1-scores but are the fastest, with inference times under 2 seconds. In the FakeNewsNet dataset, Llama3 70B achieves the highest F1-score (\textasciitilde95\%), though this comes with a high time cost similar to Gpt4-turbo (\textasciitilde1500 sec). RoBERTa again scores an excellent score of 93\% in 4 seconds, and NB and SVM achieve F1-scores of over 88\% in less than 1 second, indicating that for binary classification tasks like FakeNewsNet, traditional models can achieve competitive results, in an exceptionally short amount of time. The overall trend across both datasets shows that while larger LLMs can achieve higher F1-scores, traditional models, and smaller LLMs provide impressive results with significantly lower total response time.

\vspace*{-0.5\baselineskip}
\section{Conclusion and Future Work}
\vspace*{-0.7\baselineskip}
The results showcase that Large Language Models can match and even exceed the performance of traditional state-of-the-art methods, especially in complex classification tasks, which, however, usually comes with higher response times. Llama3 70B showed top-tier performance across both tasks and various prompting methods, while Gpt4-turbo excelled in the 3-class classification task.

We observe that the ML approaches NB and SVM can achieve excellent performance on simpler - binary classification scenarios, outperforming most of the models. However, with a more complex problem like 3-class classification task both - NB and SVM demonstrate some of the lowest performance.

While RoBERTa does not match the peak performance of larger LLMs, especially in the complex classification scenario, it achieves competitive results across both tasks with significantly shorter inference times. This highlights the trade-off between time efficiency and performance, making model selection highly dependent on the specific performance and speed requirements of a task.

The effectiveness of prompting methods varied across models and tasks, showing that task phrasing can significantly impact the results, especially in lower-performing models. We observe that the CoT technique and the FS setting are able to offer notable performance increases to the models. Both provide the models with supplementary, task-relevant information that benefits the results.

In future work, we intend to extend our experiments by including a broader selection of datasets and models, focusing on the trade-off between price and performance across the models.
Additionally, we aim to incorporate more prompting combinations to further investigate their influence on model performance. This will enable a more comprehensive examination of LLMs and allow us to discern whether the performance disparities stem from differences in model scale, inherent characteristics of the underlying LLM base model, or training datasets and methodologies. We also plan to explore additional datasets from diverse domains such as medical, legal, and social media, and  particularly datasets with higher complexity, such as datasets with longer text sequences or more than three classes.

\bibliographystyle{unsrt}  
\bibliography{templateArxiv}

\end{document}